%% file: root.tex
\documentclass[letterpaper, 10 pt, conference]{ieeeconf}  

\IEEEoverridecommandlockouts                              

\usepackage{graphics} 
\usepackage{epsfig} 
\usepackage{mathptmx} 
\usepackage{times} 
\usepackage{amsmath} 
\usepackage{amssymb}  

\usepackage{comment}
\usepackage{todonotes}
\usepackage{booktabs}
\usepackage{hyperref}
\usepackage{blindtext}

\usepackage{subcaption}
\usepackage{multirow}

\title{\LARGE \bf
Retargeting Matters: General Motion Retargeting for Humanoid Motion Tracking
}

\author{João Pedro Araújo$^{\dagger}$\quad Yanjie Ze$^{\dagger}$\quad Pei Xu$^{\dagger}$\quad Jiajun Wu$^{*}$\quad C. Karen Liu$^{*}$\\
\textbf{Stanford University}
\thanks{\textdagger Equal contribution. * Equal advising.}%
\thanks{$^{1}$João Pedro Araújo, Yanjie Ze, Pei Xu, Jiajun Wu, and C. Karen Liu are with Department of Computer Science, School of Engineering, Stanford University,%
        353 Jane Stanford Way, Stanford, CA 94305, United States \newline%
        Send correspondence to: {\tt\small jparaujo@stanford.edu}}%
}

\begin{document}

\maketitle
\thispagestyle{empty}
\pagestyle{empty}

\begin{abstract}
Humanoid motion tracking policies are central to building teleoperation pipelines and hierarchical controllers, yet they face a fundamental challenge: the embodiment gap between humans and humanoid robots. Current approaches address this gap by retargeting human motion data to humanoid embodiments and then training reinforcement learning (RL) policies to imitate these reference trajectories. However, artifacts introduced during retargeting, such as foot sliding, self-penetration, and physically infeasible motion are often left in the reference trajectories for the RL policy to correct. While prior work has demonstrated motion tracking abilities, they often require extensive reward engineering and domain randomization to succeed. In this paper, we systematically evaluate how retargeting quality affects policy performance when excessive reward tuning is suppressed. To address issues that we identify with existing retargeting methods, we propose a new retargeting method, General Motion Retargeting (GMR). We evaluate GMR alongside two open-source retargeters, PHC and ProtoMotions, as well as with a high-quality closed-source dataset from Unitree. Using BeyondMimic for policy training, we isolate retargeting effects without reward tuning. Our experiments on a diverse subset of the LAFAN1 dataset reveal that while most motions can be tracked, artifacts in retargeted data significantly reduce policy robustness, particularly for dynamic or long sequences. GMR consistently outperforms existing open-source methods in both tracking performance and faithfulness to the source motion, achieving perceptual fidelity and policy success rates close to the closed-source baseline. Website: \href{https://jaraujo98.github.io/retargeting_matters}{jaraujo98.github.io/retargeting\_matters}. Code: \href{https://github.com/YanjieZe/GMR}{github.com/YanjieZe/GMR}.

\end{abstract}

\input{sections/1_intro}
\input{sections/2_related}
\input{sections/3_methods}
\input{sections/4_experiments}
\input{sections/5_conclusion}  

\section*{ACKNOWLEDGMENT}

This work is in part supported by the Stanford Institute for Human-Centered AI (HAI), the Stanford Wu-Tsai Human Performance Alliance, ONR MURI N00014-22-1-2740, Google, and Meta Reality Labs.

\bibliographystyle{IEEEtran}
\bibliography{IEEEexample}

\end{document}

%% file: sections/1_intro.tex
\section{Introduction}

Developing humanoid policies that truly generalize to real-world environments requires learning from data that captures physical interaction with the world. Given the morphological similarities between humans and humanoids, recent work~\cite{darvish2019whole,he2024learning,cheng2024expressive,radosavovic2024real,fu2024humanplus,ze2025twist,zhang2025hub,xie2025kungfubot} has leveraged 3D human motion data (sourced from motion capture \cite{AMASS:ICCV:2019} or human motion recovery from video \cite{zhang2025hub, xie2025kungfubot, videomimic}) as demonstrations to train humanoids to perform whole-body movements requiring human-like balance and agility. These humanoid motion tracking policies are a fundamental tool for building teleoperation pipelines or hierarchical control systems. However, significant differences still exist between humans and humanoids in terms of bone length, joint range of motion, kinematic structure, body shape, mass distribution, and actuation mechanisms. This embodiment gap is the first major hurdle that must be overcome for 3D human motion data to be fully useful for humanoid learning.

The standard approach for overcoming this embodiment gap is to use kinematic retargeting from the source human motion to the target humanoid embodiment. Given the retargeted data, a popular practice in current robotics research is to use a reinforcement learning (RL) based approach to learn a policy capable of achieving a desired task through reference motion imitation. In most cases (for an exception see \cite{he2025asap}), this policy is then deployed zero-shot into the real world. This practice either overlooks glaring artifacts introduced by the retargeting process (such as foot sliding, ground penetration, and physically impossible motion due to self-penetration), instead forcing the RL policy to imitate physically infeasible motions while maintaining physical constraints, or discards the poorly retargeted data \cite{zhang2025hub}. Prior work~\cite{gu2024humanoid,he2024hover,he2024omnih2o,he2025asap} has shown that while training policies on retargeted data with severe artifacts in simulation is possible, transferring them to the real world demands extensive trial-and-error, reward shaping, and parameter tuning. Considering this practice, our hypothesis is that \emph{with enough engineering in the reward function and domain randomization, the artifacts caused by retargeting can be mostly mitigated or removed}. However, without these engineering efforts, \emph{the quality of retargeting results plays a significant role}.  

In this paper, we conduct rigorous experiments and analysis to validate our hypothesis. We compare three methods for human to humanoid motion retargeting applied to motion tracking tasks that do not involve interaction with an object or a complex scene: \textbf{PHC}, widely used by recent humanoid motion tracking works \cite{he2024learning,he2025asap}; \textbf{ProtoMotions}, used for retargeting challenging dynamic tasks \cite{xie2025kungfubot}; and \textbf{GMR}, a general motion retargeting method proposed by us to address the problems we find in the other two methods. Specifically, we find that the way in which PHC and ProtoMotions handle human to robot scaling introduces several motion artifacts that negatively impact performance. GMR addresses this by using a simple but flexible non-uniform local scaling procedure, followed by a two-stage optimization to solve for the robot motion that tracks the scaled reference.  In addition, we compare the three retargeting methods with a high-quality retargeted motion dataset generated by a closed-source method not available to the public (\textbf{Unitree}).

To isolate the impact of retargeting methods, we use BeyondMimic \cite{truong2025beyondmimic} for training and evaluation of RL policies that track given reference motions. BeyondMimic does not depend on reward tuning and is developed independently from the retargeting methods we use, making it a fair method to evaluate them.
We train single-trajectory policies for a diverse subset of the LAFAN1 dataset, excluding motions with contacts other than feet. Our final dataset consists of 21 sequences with lengths ranging from 5 seconds to 2 minutes.

For evaluation, we compare success rates measured using a very strict success definition that requires the policy to complete the reference motion in its entirety for the rollout to be considered successful. We evaluate each policy a significant number of times under conditions that take into account observation noise (both from noisy sensors and from state estimation algorithms), errors in model parameter estimation, and network latency in the controller.
We also conduct a user study to evaluate the perceptual faithfulness of the retargeted motions to the source human motion. This additionally tells us if the policies will be learning the motions we intend them to learn, or if they are learning a variant (which might be easier or harder to track).

Our end-to-end experiments demonstrate that the choice of retargeting method critically impacts humanoid performance. While policies trained on motions retargeted with different methods are generally capable of tracking a wide range of motions, including both simple and highly dynamic ones (consistent with results presented in prior work), there are a few exceptions where artifacts introduced during retargeting make it more difficult (and in some cases impossible) for the policy to learn effectively. These cases highlight that without the extensive reward engineering found in prior works, retargeting artifacts do pose challenges for certain motions and reduce policy performance. We found that foot penetration, self-intersection, and abrupt velocity spikes are all critical artifacts that should be avoided during retargeting. Additionally, we note that the initial frame of the reference motion can greatly impact whether the policy is able to start tracking it or if it fails immediately (regardless of retargeting method used), something that was observed in prior work \cite{zhang2025hub}. Finally, our results show that GMR comes very close to the Unitree retarget dataset in both faithfulness score and policy success rate, while also having higher success rates and lower motion tracking errors than the other methods. In summary, the contribution of this work includes:
\begin{itemize}
    \item A new general motion retargeting method, GMR, that addresses issues found in other retargeters and produces high-quality retargeted motion from a wide range of human motion.
    \item A comprehensive study on the impact of retargeted reference motion quality on the performance of humanoid motion tracking policies.
\end{itemize}

%% file: sections/2_related.tex
\section{Related Work}
\label{sec:related_work}

Motion retargeting is a common data processing technique for character animations in computer graphics. 
Classic methods~\cite{10.1145/311535.311536,10.1145/1037957.1037963,10.5555/1410368.1410371,803346} employ optimization-based approaches and rely on heuristically defined kinematic constraints to map motions to articulated characters. 
With the advancement of deep learning techniques,
data-driven approaches have drawn lots of attention in recent years.
These approaches, however, typically require paired data to perform supervised learning~\cite{uk2020variational,lee2023same}, 
need semantic labels to perform model training in an unsupervised manner~\cite{villegas2018neural,aberman2020skeleton,hu2023pose}, 
or use language models and differentiable rendering techniques to perform evaluation visually~\cite{zhang2024semantics}.
Besides retageting for single, rigid-body characters, previous literature also developed approaches for retargeting multiple interacting characters~\cite{liu2006composition,zhang2023simulation,jang2024geometry}, and characters with deformable shapes~\cite{liu2018surface,villegas2021contact,zhang2023skinned}.

In the robotics community, though data-driven approaches have been widely used for controlling humanoid robots~\cite{he2024learning,cheng2024expressive,radosavovic2024real,fu2024humanplus,zhang2025hub,xie2025kungfubot,ze2025twist,videomimic,truong2025beyondmimic} to generate human-like motions through imitation learning,
the difficulty of acquiring paired or semantically labeled motion data on real robots limits the application of data-driven retargeting approaches on humanoids.
While some works~\cite{stanley2021robust,yan2023imitationnet,yagi2024unsupervised} explored learning-based approaches of motion retargeting on humanoids, they focused only on simple arm and upper-body motions.
In this paper,
we focus on methods for whole-body motion retargeting involving locomotion and that do not require pre-collecting any data.

Na\"{i}ve approaches~\cite{cheng2024expressive,fu2024humanplus} directly copy the joint rotation from the source human motions to the joint space of the target humanoid.
However, the topological and morphological differences between human subjects and the humanoid often lead to artifacts like floating, feet penetrations and sliding, and the drift of end effectors (feet and hands).
Besides, additional processing is needed to convert the $SO(3)$ joint space from humans to humanoids (typically equipped with only revolute joints).

By solving the inverse kinematics~(IK) problem,
approaches of whole-body geometric retargeting~(WBGR) perform whole-body retargeting while allowing the source and target joint spaces to be misaligned.
The vanilla WBGR~\cite{penco2018robust,darvish2019whole} ignores the size difference in the Cartesian space and performs IK to only match the orientations of key links.
HumanMimic~\cite{tang2024humanmimic}, on the other hand, solves IK for Cartesian position matching of key points while using manually defined coefficients to scale the source motions.

Taking advantage of the recent work on human body representation in computer graphics, H2O \cite{he2024learning} uses the SMPL~\cite{SMPL:2015} model to fit the robot shape as a human body and then uses it to scale the motion before solving the IK problem. A reference implementation can be found in the PHC \cite{Luo2023PerpetualHC} code base, and we refer to this method as the PHC retargeting method throughout this paper. It uses the gradient descent method to solve the IK problem through forward kinematics, which is time-consuming and limits its application for real-time scenarios.
Although used in many follow-up works~\cite{he2024omnih2o,he2024hover,jiang2024harmon,he2025asap}, 
the PHC method does not take into account the contact state of motions during retargeting, which can lead to artifacts like floating, foot sliding, and penetrations with the floor.
Moreover, SMPL is designed for human-body representation, and cannot cover well robots that have relatively large morphological discrepancies with humans.
Other works \cite{ProtoMotions, xie2025kungfubot, ze2025twist} have explored the use of differential IK solvers \cite{Zakka_Mink_Python_inverse_2025}. These methods scale the Cartesian joint positions of the source motions, and then calculate generalized velocities that when integrated in-place reduce the Cartesian joint position and orientation error between the scaled source motion and the robot. The method in ProtoMotions \cite{ProtoMotions} uses global axis-aligned scaling factors to scale the Cartesian positions of the joints in the source motion, and then minimizes a weighted sum of the position and orientation errors of matching key bodies between the source human motion and the robot. KungfuBot \cite{xie2025kungfubot} uses the ProtoMotions approach, but with scaling disabled.

%% file: sections/3_methods.tex
\section{Evaluation Method}

The goal of the evaluation in this paper is to answer the following questions.

\begin{itemize}
    \item \textbf{Q1.} Does the choice of retargeting method impact the performance of the motion tracking policies?  
    \item \textbf{Q2.} What retargeting artifacts negatively impact the policy and prevent it from being able to learn?  
    \item \textbf{Q3.} To what extent do different retargeting methods preserve the ``look'' of the source motion?
\end{itemize}

To answer these questions, we retarget the data using three methods, PHC, ProtoMotions, and GMR (Section \ref{sec:GMR}), and use a dataset retargeted with a closed-source method (Unitree). We train motion tracking policies using the BeyondMimic \cite{truong2025beyondmimic} pipeline, which has been shown to work for all types of reference motions seen in prior work without the need for reward tuning or extensive domain randomization. Crucially, BeyondMimic was developed independently of the retargeting methods under study.

\subsection{Retargeting Methods}

\textbf{PHC \cite{Luo2023PerpetualHC, he2024learning, he2024omnih2o, he2025asap}.} The PHC retargeting method is designed for motions in the SMPL \cite{SMPL:2015} format. SMPL is a parametric human body model $f_{\text{SMPL}}$ that, given a vector of shape parameters $\beta^h$ and pose parameters $\theta^h$, returns the 3D locations of vertices of a posed human body mesh, $v^h = f_{\text{SMPL}}\left(\beta^h,\ \theta^h\right)$. Given the vertex coordinates, a joint regressor $J \in \mathbb{R}^{\left(n_j\times3\right)\times \left(n_v\times 3\right)}$ is used to regress the 3D positions of the joints, $j^h = Jv^h$, which are considered the origins of the key bodies. A motion in SMPL format is thus a file with the shape parameters of the body and a tensor of pose parameters where one of the dimensions corresponds to time. The first step in this retargeting method is to fit a set of shape parameters $\beta^r$ to the robot skeleton. Then, given the pose parameters of the human motion $\theta^h_t$, the SMPL body model is used to calculate the target 3D coordinates for the robot joints over time, $\mathbf{p}^{\text{target}}_t = Jf_{\text{SMPL}}\left(\beta^r,\ \theta^h_t\right)$. The second step is to optimize the robot root translation $\mathbf{p}_t$, orientation $\mathbf{R}_t$, and joint angles $\mathbf{q}_t$ at each time step to minimize the position error between the joints of the posed robot (computed through forward kinematics) and the targets computed from the SMPL robot model,

\begin{equation}
\begin{array}{rl}
\text{min}_{\mathbf{p}_t, \mathbf{R}_t, \mathbf{q}_t} & \frac{1}{T} \sum_{t=1}^T \|\mathbf{p}^{\text{target}}_t - \text{FK}\left(\mathbf{p}_t, \mathbf{R}_t, \mathbf{q_t}\right)\|_2 \\
\text{subject to} & \mathbf{q}^- \leq \mathbf{q}_t \leq \mathbf{q}^+
\end{array}.
\end{equation}
The PHC implementation uses gradient descent on the L2 norm of the error averaged over all frames to solve this optimization. The optimizer for the root pose is Adam, and the optimizer for the joint values is Adadelta. The joint limit constraint is enforced through clamping. After the optimization is done, forward kinematics is used to compute the height of all robot bodies in the retargeted motion clip, and the lowest height is subtracted from the root translation.

\textbf{ProtoMotions \cite{ProtoMotions}.} The ProtoMotions package is a collection of standard implementations of popular methods for training motion imitation policies. It comes with an optimization-based retargeting algorithm. The source motion is scaled using a custom scaling factor for each of the world frame axes. It then uses Mink \cite{Zakka_Mink_Python_inverse_2025} to minimize the joint position and orientation errors between the scaled human and robot key bodies. Like PHC, there is also a post-processing step to modify the height, but instead of setting the lowest height to $0$ it sets it to the lowest height in the source motion.

\textbf{GMR.} We also evaluate GMR, a retargeting method proposed by us. The main motivation for developing GMR was to fix glaring artifacts found in the retargets generated by the other two methods, namely deviation from the source motion, foot sliding, ground penetrations, and self-intersections. GMR is described in detail in Sec.~\ref{sec:GMR}. The main difference from the prior two methods is how it handles source motion scaling, which we found to be the cause for many of the artifacts. This is followed by a two-stage optimization to find the robot motion.

\textbf{Unitree.} Despite not having access to the retargeting procedure, the Unitree retarget of the LAFAN1 dataset has been used in recent papers such as BeyondMimic and VideoMimic \cite{videomimic}. We include it as a baseline to gauge how the open retargeting methods compete with proprietary retargeting.

\subsection{Data Processing}

We select a diverse sample from the LAFAN1 dataset \cite{harvey2020lafan1}, ranging from simple motions like walking and turning to dynamic and complex motions such as martial arts, kicks, and dancing. We do not include motions with complex interaction with the environment, such as crawling or getting up from the floor (we exceptionally included a sequence with a cartwheel since the robot either has the feet or the hands on the floor, but never both). The full list of motions can be found in Tab.~\ref{tab:sr}.  

We retarget each motion sequence to a Unitree G1 robot. LAFAN1 files are provided in the BVH format, which GMR is directly compatible with. Both PHC and ProtoMotions require the source motion data to be in the SMPL format (ProtoMotions additionaly supports the SMPL-X \cite{SMPL-X:2019} format). We handle this conversion from BVH to SMLP(-X) following a method similar to PHC retargeting. First, we fit the shape parameters $\beta$ of the SMPL(-X) body model to the BVH skeleton by minimizing the joint position error between the two skeletons (we also penalize $\|\beta\|_2$ since we find large values of it lead to large distortions of the human mesh). Then, we leverage the fact that the LAFAN1 skeleton has the same kinematic structure of SMPL(-X) and copy the matching joint 3D rotations. Finally, we calculate the root translation as the offset that minimizes the position error between the posed LAFAN1 skeleton and the posed SMPL(-X) skeleton.
We then use the PHC and ProtoMotions retargeting methods to obtain the final G1 motion. We find that the SMPL-X body model fits the LAFAN1 skeleton better than SMPL, so we use it as source for ProtoMotions.

Although the PHC retargeting code includes a post-processing step to fix foot penetrations, we find that for some sequences this yields severe (30cm or higher) floating. We fix this by running forward kinematics on the retargeted sequences, storing the minimum body height at each frame, and then offsetting the entire motion by the mean minimum body height. 
The retargets generated by the other methods do not require similar post-processing.

\subsection{Motion Tracking Evaluation}

We use BeyondMimic to train individual motion imitation policies for each retargeted motion in IsaacSim. We measure the policies' robustness to observation noise and domain shift. To this end, we evaluate each policy $100$ times with domain randomization disabled (\textbf{sim}), and $4096$ times with domain randomization enabled (\textbf{sim-dr}). The robot starts each evaluation rollout from its default pose. We run the policy until either the robot falls or the reference ends.  

The BeyondMimic release comes with a proprietary package to deploy policies to the Unitree G1 using ROS. This package handles both state estimation and inference. To validate the robustness of the policy before deployment, this package also comes with a ROS node running MuJoCo.  
We leverage this to thoroughly and safely evaluate our policies in a setup mimicking that used in the real world. We refer to this condition as \textbf{sim2sim} evaluation. For each policy, we roll it out $100$ times in this ROS environment, and record the robot state in the simulation. As in the \textbf{sim} setting, the robot starts each episode from the default pose. The sources of randomness in this setup are the timing and synchronization conditions associated with ROS, and the noise associated with the state estimation module. There is no tuning of the simulator parameters, and the controller does not have access to privileged simulator information such as the ground truth global root pose.

\subsection{Metrics}

We evaluate both the ability of the policy to maintain balance, as well as the tracking performance. For each evaluation rollout, we consider it a success if the policy is able to reach the end of the episode without the robot anchor body height or orientation deviating from the reference by more than a given threshold (upon which the episode terminates). We report the success rate, defined as the ratio between the number of successful rollouts and the total number of rollouts. Following \cite{xie2025kungfubot}, we also evaluate the tracking performance using the average position error of body parts in global coordinates ($E_{\text{g-mpbpe}}$, mm), the average position error of body parts relative to the root position ($E_{\text{mpbpe}}$, mm), and the average angular error of joint rotations ($E_{\text{mpjpe}}$, $10^{-3}$ rad). The tracking errors are computed for all frames that a policy is alive.

\subsection{User Study Evaluation}

\begin{figure}[t]
    \centering
    \includegraphics[width=0.9\linewidth]{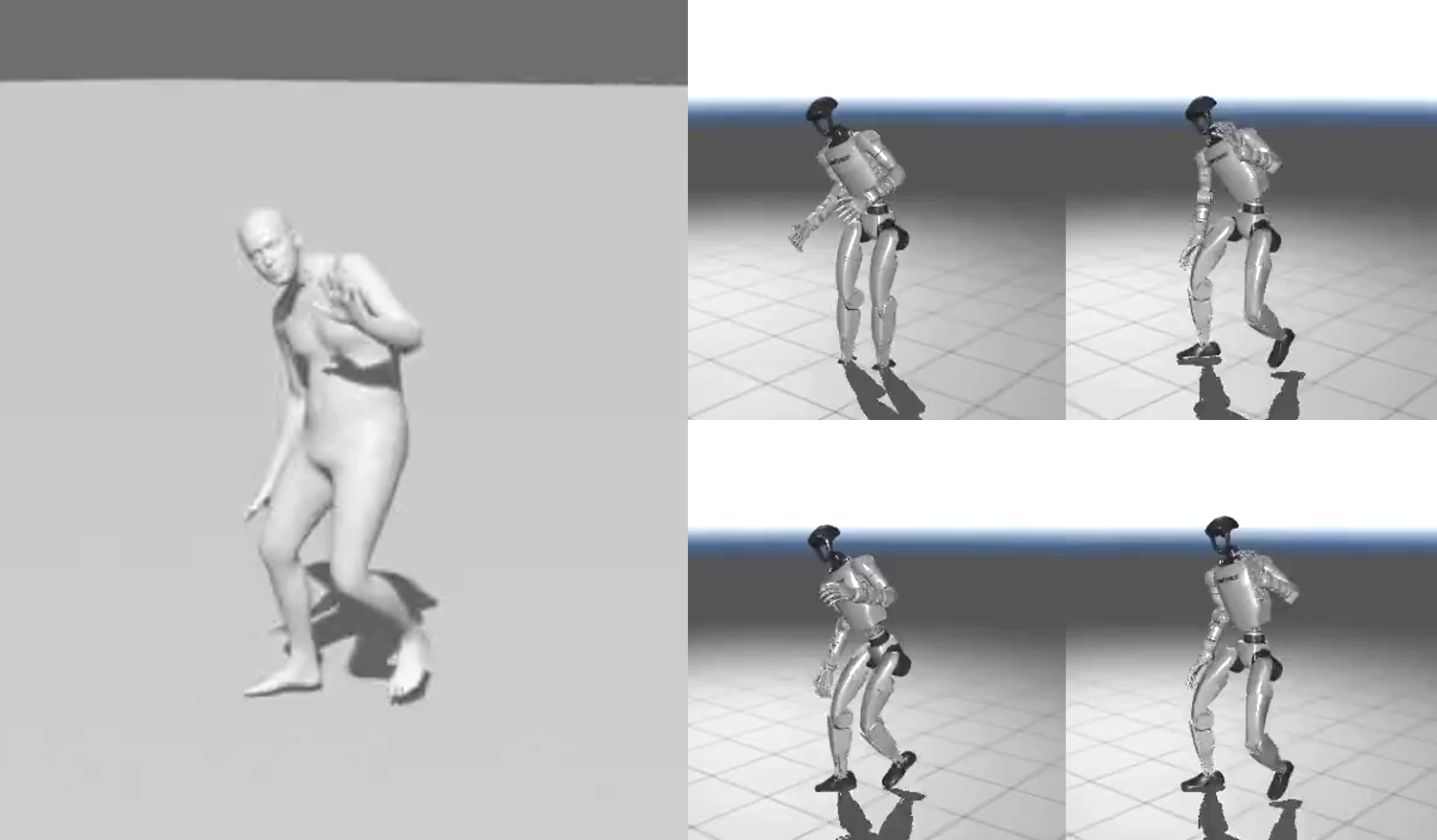}
    \begin{minipage}{0.4\linewidth}
      \centering
      \hspace{-5mm}(a) Reference motion
    \end{minipage}%
    \begin{minipage}{0.4\linewidth}
      \centering
      (b) Retarget videos
    \end{minipage}
    \caption{For the user study, participants were shown videos of the reference motion (a), and asked to choose which retarget video (b) was more similar to it.}
   \label{fig:experiments:user_study}
    \vspace{-0.15in}
    
\end{figure}

To answer \textbf{Q3}, we conduct a user study where participants are shown a $5$ second clip of a reference motion rendered using the SMPL-X fit of the LAFAN1 data, as well as two retargeted clips, one using GMR, and the other using one of Unitree, PHC, or ProtoMotions (Fig.~\ref{fig:experiments:user_study}). The users aren't told which is which and the order in which the retarget videos are presented is randomized. The users are then asked to select the video they think is closer to the reference, with the option of saying they can't find a difference. The $5$ second references are sampled randomly from $15$ of the motions. Users compare all $3$ methods to GMR for every motion, answering a total of $45$ questions.

\section{General Motion Retargeting}
\label{sec:GMR}

\begin{figure}[t]
    \centering
   \includegraphics[width=1.0\linewidth]{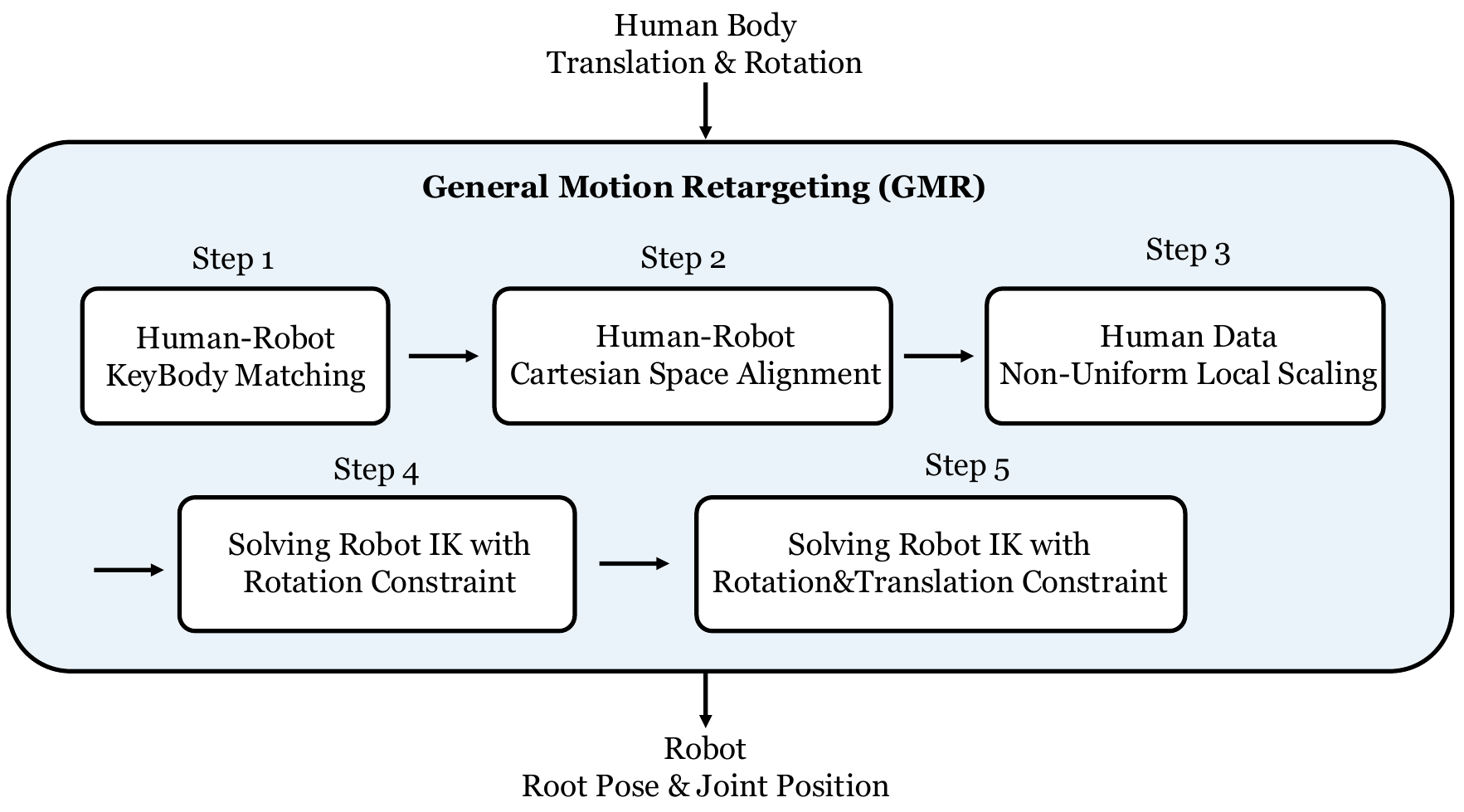}
    \caption{General Motion Retargeting (GMR) Pipeline.}
    \label{fig:gmr}
    \vspace{-0.2in}
\end{figure}

In this section we describe our proposed retargeting pipeline, General Motion Retargeting (GMR). An overview can be found in Fig.~\ref{fig:gmr}. We describe each step in detail below.

\noindent\textbf{Step 1: Human-Robot Key Body Matching.} 
Starting from a list of the bodies in the source human skeleton (which can come from a motion capture system or files in formats such as BVH and SMPL) and the target humanoid skeleton (found in XML or URDF robot description files), the user first defines the mapping $\mathcal{M}$ between the human and robot key bodies (generally torso, head, legs, feet, arms, and hands). This information is used to setup the optimization problem for the IK solver. The user can also provide weights for the position and orientation tracking errors of these key bodies.

\noindent\textbf{Step 2: Human-Robot Cartesian Space Rest Pose Alignment.} We offset the orientation of the human bodies so that they match the orientations of the robot bodies when both are in the rest pose. In some cases, we also add a local offset to the position of a body. This helps mitigate artifacts like the toed-in artifact described in \cite{he2024learning}.

\noindent\textbf{Step 3: Human Data Non-Uniform Local Scaling.} We find that most artifacts found in the other retargeting methods are introduced when scaling the source motion, highlighting how critical correct scaling is. The first step in our scaling procedure is to calculate a general scaling factor based on the height of the source human skeleton. This general factor is used to adjust a custom local scale factor defined for each key body. Having custom scale factors enables us to account for scaling differences between the lower body and the upper body, for example. The target body positions in Cartesian space are given by

\begin{equation}
    \mathbf{p}_b^{\text{target}} = \frac{h}{h_{\text{ref}}}s_b(\mathbf{p}_j^{\text{source}} - \mathbf{p}_{\text{root}}^{\text{source}}) + \frac{h}{h_{\text{ref}}}s_{\text{root}}\mathbf{p}_{\text{root}}^{\text{source}},
\end{equation}
where $h$ is the height of the human source skeleton, $h_{\text{ref}}$ is a reference height assumed when setting the scaling factors, $\mathbf{p}_b$ denotes a body position, and $s_b$ is the scaling factor corresponding to body $b$. Note that, when the body is the root, the scaling equation simplifies to

\begin{equation}
    \mathbf{p}_{\text{root}}^{\text{target}} = \frac{h}{h_{\text{ref}}}s_{\text{root}}\mathbf{p}_{\text{root}}^{\text{source}}.
\end{equation}
We find that scaling the root translation by a uniform scaling factor is crucial to avoid introducing foot sliding artifacts.

\noindent\textbf{Step 4: Solving Robot IK with Rotation Constraints.} We wish to find robot generalized coordinates $\mathbf{q}$ (root translation, root rotation, and joint values) that minimize the body position and orientation errors relative to the reference. To avoid local optimization minima, we adopt a two stage process. Given a target pose, in the first stage we solve the following optimization problem, which considers only body orientations and positions of the end-effectors,

\begin{equation}\label{eq:opt1}\begin{array}{rl}
    \min_{\mathbf{q}}&  \sum_{(i, j) \in \mathcal{M}} \left(w_1\right)^R_{i,j} \| R_i^h \ominus R_j(\mathbf{q}) \|_2^2 \\
    & + \sum_{(i, j) \in \mathcal{M}_\text{ee}} \left(w_1\right)^p_{i,j} \| \mathbf{p}_i^{\text{target}} - \mathbf{p}_j(\mathbf{q}) \|_2^2 \\
    \text{subject to}
        & \mathbf{q}^- \leq \mathbf{q} \leq \mathbf{q}^+
\end{array}
\end{equation}
In the above, $R^h_i \in SO(3)$ is the orientation of the human body $i$, $\mathbf{p}_j(\mathbf{q})$ and $R_j(\mathbf{q}) \in SO(3)$ are the Cartesian position and orientation of the robot body $j$ (obtained through forward kinematics), $R_i \ominus R_j$ is the exponential map representation of the orientation difference between $R_i$ and $R_j$ ($R_i \ominus R_j := \exp(R_i^{-1} R_j)$ in $\mathfrak{so}(3)$), $\mathcal{M}_\text{ee}$ is the subset of $\mathcal{M}$ containing only the end-effectors (hands and feet), and $\left(w_1\right)^p_{i,j}$ and $\left(w_1\right)^R_{i,j}$ are position and orientation error weights for this first optimization stage. The root position and orientation components of $\mathbf{q}$ are initialized with the scaled position $\mathbf{p}_{\text{root}}^{\text{target}}$ and yaw component of the orientation of the human root key body.
The optimization is subject to to the joint minimum and maximum values, $\mathbf{q}^-$ and $\mathbf{q}^+$. We find that sometimes this range needs to be tightened to avoid non-human movements. We solve this problem using Mink, a differential IK solver. This means that, rather than finding the values of $\mathbf{q}$ that minimize our cost function, we compute generalized velocities $\mathbf{\dot{q}}$ that when integrated reduce our cost. This is done by solving the following optimization

\begin{equation}\begin{array}{rl}
    \min_{\mathbf{\dot{q}}} & \|e(\mathbf{q}) + J(\mathbf{q})\mathbf{\dot{q}}\|_W^2\\
    \text{subject to} & \mathbf{q}^- \leq \mathbf{q} +  \mathbf{\dot{q}} \Delta t \leq \mathbf{q}^+
\end{array}
\end{equation}
where $e(\mathbf{q})$ is the loss function of Eq.~\ref{eq:opt1}, $J(\mathbf{q}) = \frac{\partial e}{\partial \mathbf{q}}$ is the Jacobian matrix of the loss relative to $q$, and $W$ is the weight matrix induced by $\left(w_*\right)^p_{i,j}$ and $\left(w_*\right)^R_{i,j}$. $\Delta t$ is a parameter of the differential IK solver and does not necessarily correspond to the time difference between the reference motion frames.

We run the solver until convergence (the change in the value function is lower than a given threshold, that we set to $0.001$) or a maximum number of iterations ($10$) is reached.

\noindent\textbf{Step 5: Fine Tuning using Rotation \& Translation Constraints.} Finally, we take the solution from the previous problem and use it as initial guess to solve

\begin{equation}\label{eq:opt2}\begin{array}{rl}
    \min_{\mathbf{q}}&  \sum_{(i, j) \in \mathcal{M}} \left(w_2\right)^R_{i,j} \| R_i^h \ominus R_j(\mathbf{q}) \|_2^2 \\
    & + \left(w_2\right)^p_{i,j} \| \mathbf{p}_i^{\text{target}} - \mathbf{p}_j(\mathbf{q}^r) \|_2^2 \\
    \text{subject to}
        & \mathbf{q}^- \leq \mathbf{q} \leq \mathbf{q}^+
\end{array}
\end{equation}
which uses a set of weights $\left(w_2\right)^p_{i,j}$ and $\left(w_2\right)^R_{i,j}$ different from the one used in the first stage, and takes into account the position of all key bodies. The termination conditions from the first optimization stage apply as well.

\noindent\textbf{Application to Motion Sequences.} The method described above is for retargeting a single pose. For retargeting a motion sequence, the method is applied sequentially to each frame, using the retarget result from the previous frame as initial guess to the optimization in \textbf{Step 4}. After a full motion has been retargeted, forward kinematics is used to get the height of all robot bodies over time. The minimum height is then subtracted from the global translation to fix height artifacts (floating or ground penetration).

%% file: sections/4_experiments.tex
\section{Evaluation Results}

\begin{table*}[!ht]
    \centering
    \caption{Evaluation success rates (\%) in IsaacSim (training simulator) without domain randomization (\textbf{sim}, $100$ trials per policy), with domain randomization (\textbf{sim-dr}, $4096$ trials per policy), and MuJoCo/ROS simulator (\textbf{sim2sim}, $100$ trials per policy). PM = ProtoMotions, U = Unitree. *See Section~\ref{sec:reference_initial_frame}}
    \label{tab:sr}
    \begin{tabular}{llllllllllllll}
    \toprule
        &  & \multicolumn{4}{c}{\textbf{sim}}& \multicolumn{4}{c}{\textbf{sim-dr}} & \multicolumn{4}{c}{\textbf{sim2sim}} \\
        \cmidrule(lr){3-6} \cmidrule(lr){7-10} \cmidrule(lr){11-14}
        \textbf{Motion} & \textbf{Length (s)}  & \textbf{PHC} & \textbf{GMR} & \textbf{PM} & \textbf{U} & \textbf{PHC} & \textbf{GMR} & \textbf{PM} & \textbf{U}& \textbf{PHC} & \textbf{GMR} & \textbf{PM} & \textbf{U} \\
        \midrule
        Walk 1 & 33  &100 &100 &100 &100& 100 & 100 & 100 & 100 & 100 & 100 & 100 & 100 \\ 
        Walk 2 & 5.5  &23 &100 &100 &100& 53.54 & 100 & 99.98 & 100 & 100*& 100*& 100*& 100*\\ 
        Turn 1 & 12.3  &93 &100 &100 &100& 87.18 & 99.98 & 99.95 & 100 & 100*& 100*& 99*& 100*\\ 
        Turn 2 & 12.3  &100 &100 &100 &100& 99.95 & 99.98 & 100 & 99.98 & 99 & 100 & 100 & 99 \\ 
        Walk (old) & 33  &100 &100 &100 &100& 100 & 100 & 100 & 100 & 100 & 100 & 100 & 100 \\ 
        Walk (army) & 13  &100 &100 &100 &100& 99.85 & 98.63 & 99.95 & 99.95 & 100 & 100 & 99 & 100 \\ 
        Hop & 13  &95 &100 &100 &100& 92.97 & 100 & 100 & 100 & 100 & 100 & 100 & 100 \\ 
        Walk (knees) & 19.58  &100 &100 &100 &100& 99.98 & 100 & 100 & 100 & 100 & 100 & 100 & 100 \\ 
        Dance 1& 118  &0 &100 &100 &99& 0 & 99.46 & 99.24 & 99.95 & 0 & 100 & 100 & 100 \\ 
        Dance 2 & 130.5  &0 &100 &100 &100& 0.02 & 99.9 & 99.88 & 99.98 & 0 & 100 & 100 & 100 \\ 
        Dance 3 & 120  &100 &100 &100 &100& 100 & 100 & 99.95 & 100 & 99 & 100 & 100 & 100 \\ 
        Dance 4 & 20  &100 &100 &100 &100& 100 & 100 & 100 & 100 & 99 & 100 & 100 & 100 \\ 
        Dance 5 & 68.4  &100 &96 &100 &100& 100 & 92.75 & 99.98 & 100 & 100 & 51 & 100 & 100 \\ 
        Run (slow) & 50  &100 &100 &100 &100& 99.19 & 99.88 & 99.95 & 99.98 & 100 & 100 & 100 & 100 \\ 
        Run & 11  &100 &100 &100 &100& 99.98 & 100 & 99.95 & 100 & 100 & 100 & 100 & 100 \\ 
        Run (stop \& go) & 37  &17 &98 &20 &100& 20.46 & 91.24 & 40.26 & 99.83 & 74 & 100 & 26 & 100 \\ 
        Hop around & 18  &100 &100 &100 &100& 100 & 100 & 100 & 100 & 100 & 100 & 100 & 100 \\ 
        Hopscotch & 10  &100 &100 &100 &100& 100 & 100 & 100 & 99.98 & 100 & 100 & 100 & 100 \\ 
        Jump and rotate & 21  &100 &100 &100 &100& 99.98 & 100 & 99.9 & 100 & 99 & 100 & 100 & 99 \\ 
        Kung fu & 8.6  &100 &100 &100 &100& 100 & 99.95 & 100 & 100 & 100 & 100 & 100 & 100 \\ 
        Various sports & 42.58  &100 &100 &100 &100& 99.98 & 99.98 & 99.95 & 100 & 100 & 100 & 100 & 99 \\ 
    \toprule
    \end{tabular}
    \vspace{-0.1in}
\end{table*}

\begin{table*}[]
    \centering
    \caption{Tracking errors for each policy measured over the $100$ evaluation rollouts in the \textbf{sim} setting. Lower values are better. Best values are \textbf{bold}, second best are \underline{underlined}.}
    \label{tab:experiment_results:tracking_errors}
    \begin{tabular}{lcccccccccccc}
        \toprule
        & \multicolumn{4}{c}{$E_{\text{g-mpbpe}}$, mm}& \multicolumn{4}{c}{$E_{\text{mpbpe}}$, mm} & \multicolumn{4}{c}{$E_{\text{mpjpe}}$, $10^{-3}$ rad} \\
                \cmidrule(lr){2-5} \cmidrule(lr){6-9} \cmidrule(lr){10-13}
                \textbf{Statistics} & \textbf{PHC} & \textbf{GMR} & \textbf{PM} & \textbf{U} & \textbf{PHC} & \textbf{GMR} & \textbf{PM} & \textbf{U}& \textbf{PHC} & \textbf{GMR} & \textbf{PM} & \textbf{U} \\
        \midrule
        Min & $71.8$ & \underline{$59.9$} & $66.0$ & $\mathbf{51.1}$ & $20.9$ & $\mathbf{18.1}$ & $24.1$ & \underline{$18.2$} & $569.5$ & \underline{$362.0$} & $499.0$ & $\mathbf{355.5}$ \\
        Median & $111.9$ & \underline{$91.2$} & $101.9$ & $\mathbf{73.4}$ & $29.9$ & \underline{$27.6$} & $30.4$ & $\mathbf{23.1}$ & $739.8$ & \underline{$546.0$} & $599.7$ & $\mathbf{467.2}$ \\
        Mean & $247.8$ & \underline{$104.1$} & $139.7$ & $\mathbf{77.2}$ & $40.2$ & \underline{$28.1$} & $33.2$ & $\mathbf{23.2}$ & $778.5$ & \underline{$561.7$} & $641.8$ & $\mathbf{483.0}$ \\
        Max & $1062.3$ & \underline{$200.0$} & $915.6$ & $\mathbf{131.4}$ & $134.4$ & \underline{$48.0$} & $107.9$ & $\mathbf{28.9}$ & $1336.1$ & \underline{$1044.8$} & $1397.9$ & $\mathbf{678.5}$ \\
        \toprule
    \end{tabular}
    \vspace{-0.25in}
\end{table*}

The success rates in the three evaluation setups (\textbf{sim}, \textbf{sim-dr}, and \textbf{sim2sim}) are reported in Tab.~\ref{tab:sr}. Statistics of the tracking error over the $21$ motions are reported in Tab.~\ref{tab:experiment_results:tracking_errors}.

\subsection{Impact of retargeting method in policy performance}

Out of the $21$ motions tested, $11$ (``Turn 2'', ``Walk (army)'', ``Walk (knees)'', ``Dance 3'', ``Dance 4'', ``Run (slow)'', ``Run'', ``Hopscoth'', ``Jump and rotate'', ``Kung fu'', ``Various sports'') achieve success rates above $98\%$ across all retargeting methods, and $3$ achieve perfect performance (``Walk 1'', ``Walk (old)'', and ``Hop around''). For the remaining $7$ motions, we see a large variation in performance across retargeting methods. Policies trained on the Unitree data achieve near-perfect performance on all motions, confirming the results from BeyondMimic. These results are closely followed by the policies trained on data retargeted using the GMR and ProtoMotions method, with a few exceptions (the ``Dance 5'' motion for GMR, and the ``Run (stop \& go)'' motion for ProtoMotions). PHC is the method with the lowest performance. We note that long horizon motions are not a challenge since the policy trained on the PHC retarget of the ``Dance 3'' motion (two minutes long) is successful in all settings. The failure to learn to track successfuly the ``Dance 1'' and ``Dance 2'' motions, as well as the drops in performance for specific GMR and ProtoMotions policies, is explained by artifacts in the retarget, which we describe in more detail in Sec.~\ref{sec:results:impact_of_artifacts}.

The success rates do not paint a full picture, as in order to be successful the policy just needs to reach the end of the episode without falling. Tab.~\ref{tab:experiment_results:tracking_errors} shows that policies trained on data retargeted using PHC or ProtoMotions can have considerable tracking errors in both global and local space, indicating the difficulty in tracking the original reference without becoming unstable.

The results presented in this section show that while it is possible to train motion tracking policies to be successful in tracking a wide variety of motions, this comes at the expense of higher tracking errors, and in some cases it is noticeably more difficult to learn a robust motion tracking policy from the retargeted data. As such, we can answer \textbf{Q1} affirmatively.

\subsection{Impact of retargeting artifacts}
\label{sec:results:impact_of_artifacts}

\begin{figure}[htbp]
  \centering

  \begin{subfigure}{0.48\linewidth}
    \centering
    \includegraphics[width=0.8\linewidth]{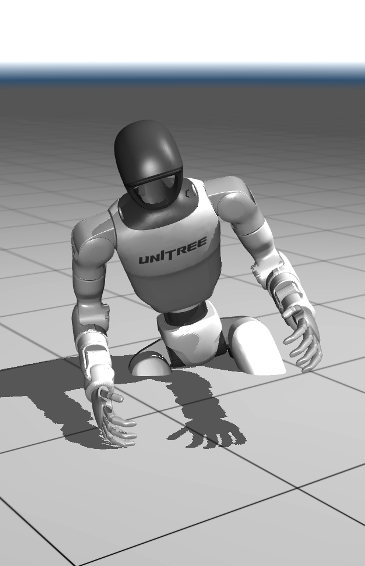}
    \caption{\textbf{Ground penetration} (PHC, ``Dance 1'')}
  \end{subfigure}%
  \hfill
  \begin{subfigure}{0.48\linewidth}
    \centering
    \includegraphics[width=0.8\linewidth]{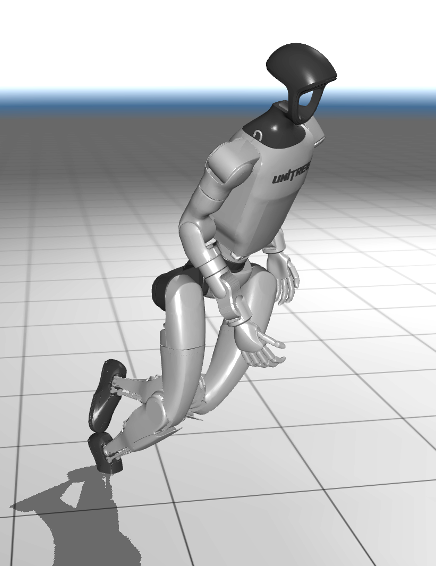}
    \caption{\textbf{Self-intersection} (ProtoMotions, ``Run (stop \& go)'')}
  \end{subfigure}

  \vskip\baselineskip
  \vspace{-0.15in}
  \begin{subfigure}{0.98\linewidth}
    \centering
    \includegraphics[width=0.9\linewidth]{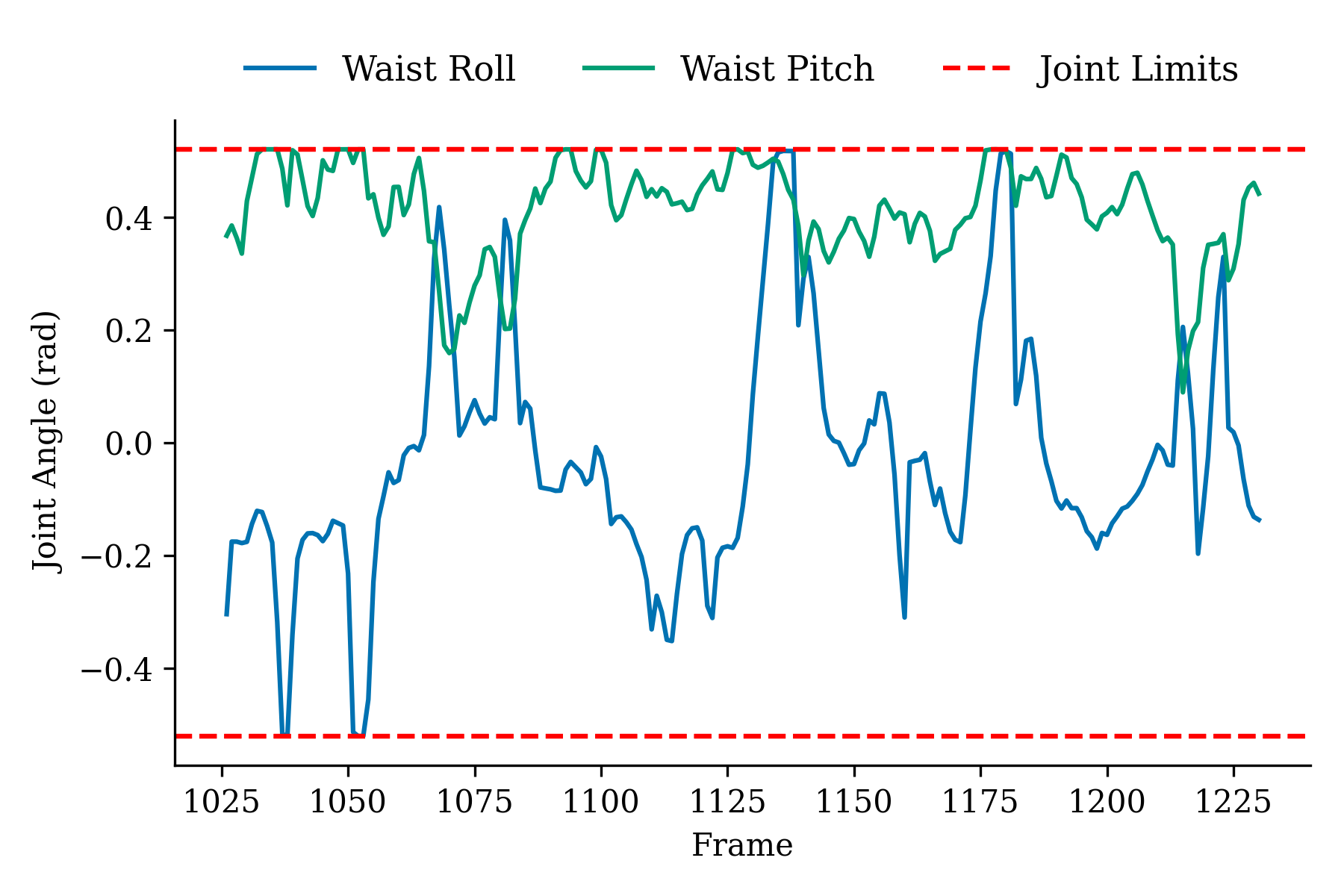}
    \vspace{-0.13in}
    \caption{\textbf{Sudden jumps} in the waist roll and pitch values (GMR, ``Dance 5'')}
  \end{subfigure}
    \vspace{-0.06in}

  \caption{Example artifacts found in the retargeted references with low success rates.}
 \label{fig:retarget_artifacts}
    \vspace{-0.2in}
  
\end{figure}

To answer \textbf{Q2}, we note that low scores in the \textbf{sim2sim} evaluation can be directly connected to artifacts in the retargeted motion (Fig.~\ref{fig:retarget_artifacts}).

\begin{itemize}
    \item The PHC retarget of the ``Dance 1'' and ``Dance 2'' motions have noticeable \textbf{ground penetration} artifacts ($60$ cm in one case).
    \item The robot legs in the ProtoMotions retarget of the ``Run (stop \& go)'' motion \textbf{intersect} with each other.
    \item The GMR retarget of the ``Dance 5'' motion has many \textbf{sudden jumps} of the waist roll value.
\end{itemize}
While their presence does not make a reference motion impossible to track (with the exception of the long dance sequences retargeted with the PHC method), these three retargeting artifact types (physically inconsistent height, self-intersections, and sudden jumps in joint values) should be avoided to ensure the best chances of success.

The sudden jumps in the GMR retargets are a rare occurrence (for the subclip of the ``Dance 5'' motion it happens on a $10$ second segment, less than 2\% of our full motion dataset) that is introduced during the optimization phase. Since we use the same optimization weights for all experiments, some motions might require further weight tuning to achieve optimal results.

\subsection{User study}

\begin{figure}[t]
    \centering
    \includegraphics[width=0.9\linewidth]{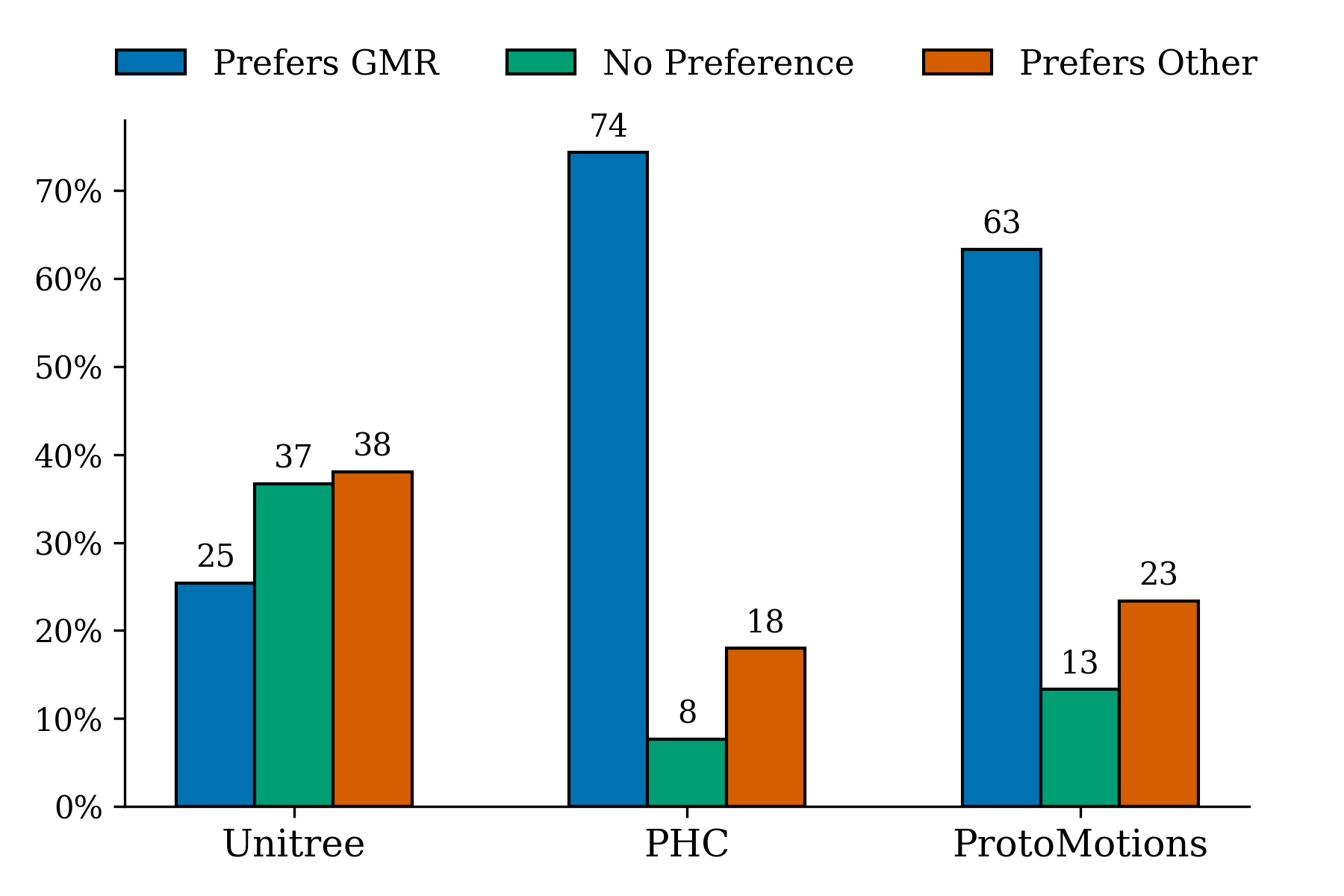}
    \vspace{-0.15in}
    \caption{User study ($N=20$) results for comparing GMR to other retargets in terms of faithfulness to the source motion. The bars represent the percentage of responses.}\label{fig:user_study_results}
    \vspace{-0.15in}
    
\end{figure}

$20$ users participated in the user study. The results are shown in Fig.~\ref{fig:user_study_results}. Users consider GMR to be more faithful to the reference motion than the retargets generated by either PHC or ProtoMotions. The Unitree retarget is considered more faithful than the GMR one, but the users also have a harder time distinguishing the two. The combination of the high success rates and the close faithfulness score shows that GMR is a viable alternative to the data retargeted by Unitree.

\subsection{First reference frame}
\label{sec:reference_initial_frame}

\begin{table}[!ht]
    \centering
    \caption{Evaluation success rates (\%) in \textbf{sim2sim} (MuJoCo) as a function of the start frame of the reference motion. PM = ProtoMotions, U = Unitree.}
    \label{tab:sr_sim2sim_first_frame}
    \vspace{1mm}
    \begin{tabular}{lrllll}
    \toprule
        \textbf{Motion}  &\textbf{Start frame}& \textbf{PHC} & \textbf{GMR} & \textbf{PM} & \textbf{U} \\
        \midrule
        \multirow{2}{*}{Walk 2}  &0& 100 & 64& 100 & 100 \\
  &7& 100 & 100 & 100 &100 \\ 
  \midrule
        \multirow{2}{*}{Turn 1}  &0& 14 & 100 & 86 & 47 \\
 & 49 & 100 & 100 & 99 & 100\\ 
    \toprule
    \end{tabular}
    \vspace{-0.2in}
    
\end{table}

Finally, we note (as has been done previously by \cite{zhang2025hub}) that the starting frame of the reference motion can have a large impact on the policy performance. In Tab.~\ref{tab:sr_sim2sim_first_frame} we show the success rates in the \textbf{sim2sim} (MuJoCo) setting for the same policy but different start frames. 
We recommend ensuring that the start pose of the reference motion is such that the robot can safely reach it once policy inference starts. Likewise, we recommend ending the reference motion on a stable pose, to allow safe deactivation of the robot.

%% file: sections/5_conclusion.tex
\section{Conclusion}

In this paper we evaluated the impact of motion retargeting on the performance of motion tracking policies. Our analysis covered GMR, a retargeting method we introduce, the popular PHC retargeting algorithm, the ProtoMotions retargeting algorithm, and the data officially retargeted by Unitree. We retargeted a diverse set of motions ranging in length and difficulty, and used BeyondMimic to train motion tracking policies for each of the retargeted clips. We thoroughly evaluated each policy to measure it's robustness to observation noise, model mismatch, and network latency, as well as it's tracking performance.

We find that while it is possible to train successful policies for a wide variety of motions retargeted using all the retargeting methods under study, there are some critical artifacts that greatly increase the difficulty in learning a motion tracking policy, namely ground penetrations, self-intersections, and sudden jumps in joint values. In addition, our user study shows that both the PHC and the ProtoMotions retargeting methods yield motions that are less faithful to the source material than the ones yielded by GMR. While in rare occasions GMR may still require some tuning to eliminate optimization artifacts, we find that the default parameters work well for a wide range of motions, yielding both good reference motions for training policies while staying faithful to the source motion, which cannot be said of the other retargeters.

There are some limitations to this work. Despite the variety of motions considered, they all came from a single source (the LAFAN1 dataset). Further study should be done with more data sources, such as the AMASS dataset or human motion reconstructed from monocular video. Another limitation is that we only consider the Unitree G1. This is a limitation primarily imposed by the BeyondMimic code base, as the other retargeting methods are available for the Unitree H1 as well. However, both BeyondMimic and retargeting algorithms benchmarked are general, and so extending this analysis to other humanoid robots should also be considered in future work. Finally, another future work direction is the impact of retargeting on motion sequences that involve interactions, for example, with the surrounding environment, with objects, or with other robots.